\documentclass{article}
\usepackage[utf8]{inputenc}

\usepackage{fullpage}

\usepackage{array}
\usepackage{caption}
\usepackage{subcaption}
\usepackage{graphicx}
\usepackage{multirow}
\usepackage{amsfonts}
\usepackage{xcolor}
\usepackage[toc]{appendix}
\usepackage[rightcaption]{sidecap}
\usepackage{amsmath}
\usepackage{authblk}
\usepackage{hyperref}
\usepackage{tikz}
\usetikzlibrary{mindmap}
\usepackage{bigints}
\usetikzlibrary{shapes, snakes, arrows}
\usetikzlibrary{positioning,fit,calc} % used for the efficient working of the positioning system  
\usepackage{pifont}% http://ctan.org/pkg/pifont
\newcommand{\cmark}{\ding{51}}%
\newcommand{\xmark}{\ding{55}}%

\tikzset{block/.style={draw, thick, text width=3cm, minimum height=1cm, align=center},   
line/.style={-latex}   % the lesser the width the greater will be the diagram window  
}  

\tikzset{elips/.style={ellipse,draw,minimum width=4em,minimum height=1cm, align=center, inner ysep=0pt}}

\newcommand\independent{\protect\mathpalette{\protect\independenT}{\perp}}
\def\independenT#1#2{\mathrel{\rlap{$#1#2$}\mkern2mu{#1#2}}}

\DeclareMathOperator*{\argmin}{arg\,min}

\usepackage[
    backend=biber,
    style=numeric,
    hyperref = true,
    natbib = true
  ]{biblatex}

\title{A Survey on Preserving Fairness Guarantees in Changing Environments}

\author[1,2]{Ainhize Barrainkua}
\author[1,3]{Paula Gordaliza}
\author[1,4]{Jose A. Lozano}
\author[1,2,5]{Novi Quadrianto}

\affil[1]{Basque Center for Applied Mathematics (BCAM), Spain}
\affil[2]{Predictive Analytics Lab (PAL), University of Sussex, UK}
\affil[3]{Institute of Mathematics of the University of Valladolid (IMUVa), Spain}
\affil[4]{University of the Basque Country UPV/EHU, Spain}
\affil[5]{Monash University, Indonesia}

\begin{document}

\maketitle

\begin{abstract}
  Human lives are increasingly being affected by the outcomes of automated decision-making systems and it is essential for the latter to be, not only accurate, but also fair. The literature of algorithmic fairness has grown considerably over the last decade, where most of the approaches are evaluated under the strong assumption that the train and test samples are independently and identically drawn from the same underlying distribution. However, in practice, dissimilarity between the training and deployment environments exists, which compromises the performance of the decision-making algorithm as well as its fairness guarantees in the deployment data. There is an emergent research line that studies how to preserve fairness guarantees when the data generating processes differ between the source (train) and target (test) domains, which is growing remarkably. With this survey, we aim to provide a wide and unifying overview on the topic. For such purpose, we propose a taxonomy of the existing approaches for fair classification under distribution shift, highlight benchmarking alternatives, point out the relation with other similar research fields and eventually, identify future venues of research.  
\end{abstract}

\section{Introduction}

Machine Learning (ML) systems are increasingly being used for high-stake decision making in applications where human outcomes are involved, ranging from healthcare \cite{seyyed2020chexclusion, potash2015predictive}, education \cite{hutchinson201950}, recruitment \cite{datta2015automated, hu2018short}, bank loan provision \cite{khandani2010consumer, hardt2016, fuster2022predictably}, to criminal incarceration \cite{chould2017, angwin2016}. Several decision-making approaches have been found to reproduce social discriminating behaviours: the effects of their predictions are disproportionate for the different population subgroups described by sensitive attributes (e.g. gender, race, age). Therefore, given their social impact, it is crucial for these systems to behave fairly. This has been one of the major concerns of the research community recently \cite{zhang2021ethics}, which has come up with a diverse set of fairness metrics and massive pool of alternatives to ensure fair algorithmic behaviours on different ML tasks. 

The evaluation of these models is mostly performed under single domain assumptions. That is, settings where the training and testing data are independently and identically drawn (\textit{i.i.d.}) from the same distribution are usually considered. Nonetheless, the above assumption is quite strong and is not always warranted \cite{zhuang2020comprehensive}. In practice, when these approaches are deployed in real-world settings, the deployment environments may and do systematically differ from the training environment. This phenomenon is known as \textit{distribution shift} and might arise due to a variety of factors, such as, the non-stationarity of the data under study \cite{ding2021}, which is very common in geographic and temporal data. For instance, in the European Union, the non-native population has significantly changed in the last years due to refugee crisis and internal migration. Additionally, access to the data is not generally feasible from all possible test environments at training time, including the cases when the target distributions correspond to future situations. For example, in natural language processing applications \cite{brunet2019, bolukbasi2016, jentzsch2019, zhao2018} the possible contexts in which the system may be deployed are infinite, and it is unfeasible to account for all of them at training time.

The dissimilarity between training and deployment environments can significantly deteriorate and potentially cause harm in fairness-critical applications \cite{chen2022fairness, ding2021, giguere2021fairness}. In the cases where distribution shift occurs, fairness requirements are satisfied on training samples, while they may be severely violated in the deployment data \cite{kallus2018residual, lan2017discriminatory}. For example, in real-world health applications, a disease predicting model that behaves fairly in a hospital according to given fairness criteria, may behave unfairly when deployed in a hospital of a different community. In the light of this new potential source of unfairness, accounting for potential distribution shifts in the context of fair machine learning is crucial: providing robust fairness guarantees under conditions of distributional shift is necessary for the safe deployment of fair decision-making models. 

Several works have been recently proposed aiming to overcome such adversity and the field is notably increasing. However, the characterisation of the field\textquotesingle s architecture is still lacking. By means of this survey, we aim to provide a unifying framework and rich overview on this emerging research area through taxonomy analysis, comparison, benchmarking proposals and connections to related research fields. For the proper characterisation of this survey, we have performed an exhaustive search in services including Google Scholar, dblp, ACM Library, IEEE Explore, Science direct and Springer link. Besides, we have examined the references of all the papers found and collected in those platforms, increasing the number of relevant papers on the field. The first pool of candidate papers has been narrowed to the ones we believe are the most representative to provide a good insight of the field. 

The reminder of this paper is organised as follows. Section \ref{s:theoretical} briefly introduces the reader into the theoretical framework this survey is grounded in, specifying the notation, the main definitions and stating the motivating problem. A clear classification of the different kinds of distribution shifts is also presented. Then, section \ref{s:similarities} highlights the parallelisms between the field of distribution shift and fairness and discusses whether specific approaches that account for both are necessary. Our main contribution is provided in section \ref{s:main}, which is a taxonomy-based overview of the research field. After, section \ref{s:datasets} describes the existing datasets and benchmarking frameworks and section \ref{s:relatedfields} draws connections with other related research areas. Finally, in section \ref{s:discussion} we discuss the interest of the present survey, noticing some of the current real-world applications and highlighting the future venues on the field.

\section{Theoretical Background}\label{s:theoretical}

\subsection{Notation}

This section describes the notation adopted throughout this survey. A \textit{domain} $\mathcal{D}$ is characterised by a feature space $\mathcal{X} \subseteq \mathbb{R}^d$, a sensitive attribute space $\mathcal{A}$, an output or label space $\mathcal{Y}$ and an associated joint probability distribution $\mathbb{P}(x,a,y)$ over the feature-label space triplet $\mathcal{X} \times \mathcal{A} \times \mathcal{Y}$; that is, $\mathcal{D} = \{ \mathcal{X}, \mathcal{A}, \mathcal{Y}, \mathbb{P}(x,a,y) \}$. Each instance is described by three random variables: $X \subset \mathcal{X}$ non-sensitive features,  $A \subset \mathcal{A}$ the sensitive attribute and  $Y \subset \mathcal{Y}$ the class label. In particular, we consider two domains: a source domain $\mathcal{D}_s$ and a target domain $\mathcal{D}_t$. The source domain will represent the population in which the original classifier has been trained and the target domain the population of the environment in which the classifier is about to be deployed. In practice, we assume to have access to a set of $n$ instances from the source domain defined by feature-label triplets $\{ (x_i, a_i, y_i) \}_{i=1}^n$ \textit{i.i.d.} from $\mathbb{P}_s(x,a,y)$. With regards to the target domain, we can either have access to $m$ labelled samples with or without the sensitive information, namely $\{ (x_i, a_i, y_i) \}_{i=1}^m$ or $\{ (x_i, y_i) \}_{i=1}^m$, $m$ unlabelled samples with or without the sensitive information, $\{( x_i, a_i ) \}_{i=1}^m$ and $\{ x_i \}_{i=1}^m$ respectively; or we might even have no information at all. The available target samples are assumed to be  \textit{i.i.d.} from $\mathbb{P}_t(x,a,y)$. 

The principal aim of the surveyed classifiers is to find the optimal parameter setting $\theta \in \Theta$ for the mapping function $f: \mathcal{X} \times \mathcal{A} \longrightarrow \mathcal{Y}$ that will most accurately and fairly describe the real data generating process of the deployment (target) environment. Since most of the literature in the field of algorithmic fairness adopts a binary classification task, we will stick to this framework for sake of clarity. The optimisation problem will be defined by a suitable loss function $\mathcal{L}(\theta,X,A,Y)$ and in some cases a set of constraints $\textbf{g}(\theta, X, A, Y)$.

The described notation is summed up in Table \ref{tab:notation}.  

\begin{table}[ht!]
    \centering
    \begin{tabular}{c|c}
        \textbf{Notation} & \textbf{Description} \\
        \hline
        $(x,a)$ & Instance \\
        $y$ & Label \\
        $\mathcal{X}, \mathcal{A}, \mathcal{Y} $ & Feature / Sensitive feature / Label space  \\
        $\mathcal{D}$ & Domain  \\
        $\mathbb{P}(\cdot)$ & Probability distribution \\
        $\mathbb{E}[\cdot]$ & Expectation  \\
        $\mathcal{L}(\cdot)$ & Loss function \\
        $\textbf{g}(\cdot)$ & Constraints \\
        $f(\cdot)$ & Predictive function  \\
        $\theta$ &  model parameter \\
        $\Theta$ & parameter space \\
        $s,t$ & subscripts denoting the source/target domain \\
        $n,m$ & dataset size of the source/target domain \\
        \hline
    \end{tabular}
    \caption{The notation adopted in this survey.}
    \label{tab:notation}
\end{table}

%When the source and target domains are related but are characterized by a different underlying distribution, the predictive performance and fairness guarantees of the classifiers that are trained on the source data deteriorate when it is deployed on the data belonging to the target domain. \\

\subsection{Algorithmic Fairness}

The Article 21 of the EU Charter of Fundamental Rights dictates that ``[a]ny discrimination based on any ground such as sex, race, colour, ethnic or social origin, genetic features, language, religion or belief, political or any other opinion, membership of a national minority, property, birth, disability, age or sexual orientation shall be prohibited.'' They describe two different discrimination scenarios: (1) \textit{direct discrimination} (disparate treatment in the US law): intentionally providing an adverse treatment based on the sensitive attribute; and (2) \textit{indirect discrimination} (disparate impact): when a seemingly ''neutral provision, criterion or practice'' disproportionately disadvantages members of a given sensitive group compared to others. This last type of discrimination highlights the systematic and structural unfairness found in the society. Nonetheless, it cannot be measured at individual level since it requires a comparison between different groups \cite{wachter2021fairness}. 

%The European Union has recently released their ”Ethics guidelines for trustworthy AI” report where it is stated that unfairness and biases must be avoided. %https://digital-strategy.ec.europa.eu/en/library/ethics-guidelines-trustworthy-ai 
%Direct discrimination can thus be proven at an individual level through explicit reference to a protected characteristic in the contested rule, whereas indirect discrimination requires group-level comparison

\subsubsection{Measuring algorithmic fairness}

A long list of definitions have been proposed to measure the unfair behaviour of algorithms (refer to \cite{verma2018fairness} for an extensive review), which are closely related to legal notions for discrimination. There are three general groups of fairness metrics: individual notions, group (or statistical) criteria and mini-max fairness. The former aim to ensure that similar individuals are treated similarly by the algorithm; the second seeks for approximate parity of one (or many) performance metric(s) across groups; lastly, the third group of fairness notions aims to improve the predictive performance of a given classifier on the sensitive group for which the worst performance was recorded.  

Formally, the notion of \textit{individual fairness} \cite{dwork2012fairness} defines a similarity metric $d$ with respect to the particular context under consideration, and requires that instances that are similar according to such metric receive similar treatment (i.e. outcome) by the algorithm:
\begin{equation}
    |P(\hat{y}_i= y| x_i, a_i) - P(\hat{y}_j= y| x_j, a_j)| \leq \epsilon \; \; \textrm{when} \; d((x_i, a_i),(x_j, a_j)) \approx 0
\end{equation}
The main drawback of individual notions is that they are subject to many strong assumptions, such as, the knowledge about the structure of the similarity metric, which is actually non-trivial. Furthermore, generating all possible data for testing this criterion is computationally infeasible. 

On the other hand, there are \textit{statistical} or \textit{group notions} of fairness, which are the most common criteria in the literature regarding algorithmic fairness since they are easier to measure and implement. Group notions of fairness study the gap of one (or multiple) performance metric(s) between different sensitive groups. The most popular statistical definitions are \textit{demographic parity}, \textit{equality of opportunity} and \textit{equalized odds}.

\textit{Demographic parity} (DP), also known as \textit{statistical parity} (SP), requires that the probability of having a positive outcome be similar across sensitive groups, and is closely related to the legal definition of \textit{indirect discrimination}. In other words, this notion claims that among the instances having the desired and undesired outcomes, the proportion of people with given sensitive attribute should be the same and therefore, match the demographic statistics across the population as a whole:
\begin{equation}
    |P(\hat{Y}= 1 | A = a) - P(\hat{Y}= 1 | A = a')| \leq \epsilon 
\end{equation}
This fairness definition is troublesome in what models might assign positive outcomes to the instances of the unprivileged group at random in order to satisfy the criterion, without any guarantees on whether those that really deserve it are being considered. Besides, it is also controversial in situations where the base rate (i.e. prevalence) significantly differ among different sensitive groups. 

In order to overcome such issue, several researchers proposed different criteria that account for the true class label in their definition, assuming that class labels are free from biases. \textit{Equality of opportunity} (EOp) requires the \textit{true positive rate} (TPR) to be similar across sensitive groups. In other words, it requires that the probability of having a positive outcome when the true label is positive to be similar across different subgroups:
\begin{equation}
    |P(\hat{Y}= 1 | A = a, Y = 1) - P(\hat{Y}= 1 | A = a', Y = 1)| \leq \epsilon
\end{equation} 

Further, \textit{equalized odds} (EO) requires approximate parity not only on the TPR but also on the \textit{false positive rates} (FPR):
\begin{equation}
    |P(\hat{Y}= 1 | A = a, Y = y) - P(\hat{Y}= 1 | A = a', Y = y)| \leq \epsilon \; \; \forall y
\end{equation}

In many applications, enforcing statistical notions of fairness has been found to compromise the predictive accuracy of the classifier \cite{corbett2017algorithmic, lipton2018does, menon2018}. Nonetheless, sacrificing accuracy is often unacceptable, specially in critical settings, such as, healthcare, where the no-harm principle \cite{ustun2019fairness} must be satisfied. In such scenarios, models are built with the aim of achieving good performance across different predefined sensitive groups \cite{diana2021minimax}: for that purpose, they are trained to minimise the error of the worst group (i.e., minimise the error of the group with the highest error), as originally done by Rawls in \cite{rawls2001justice, rawls2009theory}. 

%\novi{You should connect the technical definitions with the legal non-discrimination framework: e.g. ``Why fairness cannot be automated: Bridging the gap between EU non-discrimination law and AI'', ``On the Legal Compatibility of Fairness Definitions'', \url{https://apice.unibo.it/xwiki/bin/view/Projects/Compulaw}, etc.}.

\subsubsection{Fairness-enhancing methods}

The research area of algorithmic fairness has come up with a massive pool of alternatives to enforce fairness in different ML tasks \cite{pessach2022review}. Those approaches are usually grouped into three categories subject to the point in which the fairness-enhancing intervention is performed throughout the task: they are classified into pre-processing, in-processing and post-processing methods. \textit{Pre-processing} methods \cite{zemel2013,feldman2015,louizos2015,calmon2017,QuaShaTho19,KehBarThoQua20} transform the data to obtain an unbiased dataset to feed a standard ML classifier, whose output is expected to be fair. Preliminary works (e.g. \cite{kamiran2012data}) proposed to change the labels of some instances (massaging) or reweighing them to obtain fair datasets. Nonetheless, the newest approaches bring forward the idea of learning a new and fair representation of the data \cite{zemel2013}. \textit{In-processing} methods \cite{agarwal2018, bechavod2017, calders2010, goh2016, kamishima2012, woodworth2017, zafar2017, zafar2017b, QuaSha17} modify the algorithm itself to fulfil fairness criteria in training time. The latter is usually done by adding a regularisation term to the objective function that penalises unfairness (e.g. \cite{kamishima2012}) or by considering additional constraints on the optimisation problem that account for fairness requirements (e.g. \cite{zafar2017, woodworth2017}).  \textit{Post-processing} approaches \cite{corbett2017algorithmic, dwork2018, hardt2016, menon2018} adjust the output of a classifier so that the final performance is fair. For example, in \cite{hardt2016} the authors suggest a technique for flipping some the classifier\textquotesingle s predictions so that the final decisions satisfy fairness requirements.

\subsection{Changing Environments} \label{subs:changingenvironments}

Most of the fairness-enhancing interventions assume that the ML algorithm will be deployed in a population that is identical to the one the classifier has been trained on. 
However, such ideal conditions are hardly held in real-life applications. 
In fact, there usually exists a \textit{distribution-shift} between the training and target data, i.e., $\mathbb{P}_s(x,a,y) \neq \mathbb{P}_t(x,a,y)$. This can happen either because the classifier will be deployed in a different population, or because the source population itself is evolving. 

Consider a joint probability distribution consisting of three factors $U$, $V$ and $W$; this joint probability can be decomposed as follows:
%The joint probability distribution composed of three factors $U$, $V$ and $W$ can be decomposed as follows according to the chain rule: 
\begin{equation}
    \mathbb{P}(U,V,W) = \mathbb{P}(W|U,V) \mathbb{P}(V|U) \mathbb{P}(U).
\end{equation}
The factors are defined as: $U \subseteq \{ X,A,Y \}$, $V \subseteq \{ X,A,Y \} \backslash U$ and $W \subseteq \{ X,A,Y \} \backslash (U \cup V)$. A shift in the distribution can be caused by several different factors of this decomposition, which characterise the type of distribution-shift. 
First, we can have changes in the marginal distribution
of the features or the class label and, second, changes in the conditional distributions. Depending on which is the marginal whose distribution changes, the shift is named accordingly. 
Whereas, all the changes on the conditional distributions are named \textit{concept shift}. That is, we may have the following distribution shifts: 
\begin{itemize}
    \item If the shift arises in the marginal distribution of some attribute, that is, if there exists $U \neq \emptyset$ for which $\mathbb{P}_s(U) \neq \mathbb{P}_t(U)$, the distribution shift is named according to the characterisation of $U$:
    \begin{itemize}
        \item $(U=A)$ \textit{Demographic shift} \cite{biswas2021ensuring} arises when the distribution of the population w.r.t. the sensitive attribute varies, i.e., $\mathbb{P}_s(A) \neq \mathbb{P}_t(A)$. This shift emerges when e.g. the proportion of female individuals varies from the source to the target population.
        \item  $(U=X)$ \textit{Covariate-shift} \cite{rezaei2021robust} is a special scenario in which the source and target domains have different distributions with respect to the features $\mathbb{P}_s(X) \neq \mathbb{P}_t(X)$, but the functional form remains unaltered across domains, i.e, $\mathbb{P}_s(Y|X,A) = \mathbb{P}_t(Y|X,A)$. For example, when, due to economic trends, the socioeconomic situation of a given population evolves over time, there is covariate-shift between the distributions of the populations of subsequent years.
        \item $(U=Y)$ \textit{Label shift } (or \textit{prior probability shift}) \cite{dai2020label} comes off when the distribution of the class label $Y$ (also known as prevalence) varies from the source to the target domain, i.e., $\mathbb{P}_s(Y) \neq \mathbb{P}_t(Y)$. Furthermore, label shift also implies that the distribution of the features conditioned on the class label and sensitive attribute is preserved across domains, i.e., $\mathbb{P}_s(X|Y,A) = \mathbb{P}_t(X|Y,A)$. Label shift pops up when, e.g. the prevalence of a given disease varies for different geographical regions. 
        \item $U$ can also refer to any combination between $X$, $A$ and $Y$. In those cases, the shift is denoted as \textit{compound shift}. For instance, we can have a compound of covariate-shift and demographic shift, where $U=\{X,A\}$, $V=Y$ and $W=\emptyset$. $\mathbb{P}(V|U)$ will represent the functional form $\mathbb{P}(Y|X,A)$, which is assumed to remain unaltered across domains. Furthermore, a \textit{general distribution shift} scenario arises if $U=\{X,A,Y\}$ and, therefore, $W,V= \emptyset$. In fact, the majority of the shifts that arise in real world settings are compound shifts.  
    \end{itemize}
    \item If there exists $U \neq \emptyset$ for which $\mathbb{P}_s(U) = \mathbb{P}_t(U)$, and at least $V \neq \emptyset$ for which $\mathbb{P}_s(V|U) \neq \mathbb{P}_t(V|U)$  or $W \neq \emptyset$ for which $\mathbb{P}_s(W|V,U) \neq \mathbb{P}_t(W|U,V)$, it is a case of \textit{concept shift}. Originally concept shift referred to the situation in which a changing context induces changes in the target concepts, i.e. when $\mathbb{P}_s(Y|X,A) \neq \mathbb{P}_t(Y|X,A)$. However, it was later adopted to denote changes in other conditional distributions \cite{moreno2012unifying}. Concept shifts are very common in online learning scenarios, particularly, the case where $\mathbb{P}(Y|X,A)$ varies. For example, in a recommender system, a user\textquotesingle s interests may fluctuate due to a variety of reasons, such as, personal needs, current trends, employment status and age, which are not captured by the model, while the attributes that are actually accounted for ($\mathbb{P}(X,A)$ or $\mathbb{P}(X)$) remain unaltered. Furthermore, in healthcare applications, the progression of sick patients might vary in response to the applied medication or variations in the natural resistance of the patients.
\end{itemize}

\section{Similarities Between Distribution-Shift and Fairness}\label{s:similarities}

%\subsection{Do we need specific methods to address distribution-shift in fairness-aware scenarios?}

Over the last years, distribution-shift and fairness have been two of the well-researched areas in the field of ML and very similar methods have been proposed to address each of them. For instance, the literature of both research fields contains a relevant amount of works that focus on learning a new representation of the data that is invariant to a given change in the context: on the one hand, methods regarding distribution shift aim to learn representations that are invariant across domains, while in fairness the invariance property of the new representation is defined w.r.t. the sensitive information. Besides, methods that were explicitly developed to overcome distribution-shift have been found to perform well to address fairness-related concerns \cite{adragna2020fairness, yurochkin2020training, wang2020robust}, and vice-versa \cite{creager2020exchanging, veitch2021counterfactual}, showing that both research areas share similar general characteristics. However, the picture is not so clear when both challenges appear at the same time: Do we need specific methods to address distribution-shift in fairness-aware scenarios?

The naïve approach to tackle this problem is to employ conventional distribution-shift methods in combination with fairness-enhancing methods, addressing each issue independently. In principle, this can be done with the pre- and post-processing fairness-enhancing methods, for they can be employed in conjunction with any ML classifier. In a shifted scenario, post-processing methods could modify the predictions of any popular classifier that resolves distribution-shift to make them satisfy fairness requirements (e.g. as it is done in \cite{kallus2018residual}). In a similar manner, pre-processing methods would modify a dataset so that fairness requirements are satisfied and then, feed it to a conventional classifier that accounts for distribution-shift (e.g. \cite{iosifidis2020fabboo}). However, addressing the concerns related to fairness and distribution shift in a disjoint fashion might lead to sub-optimal solutions \cite{an2022transferring}. Therefore, it is preferable to tackle them jointly to preserve fairness guarantees and high predictive performance in the shifted environment, by modifying the conventional pre-, in- and post-processing fairness-enhancing approaches drawing inspiration from conventional shift-aware classifiers (or vice versa). For instance, many pre-processing methods aim to learn a new representation of the data that will be fair for the different sensitive groups: in order to account for distribution-shift, such methods could be modified to make the new representation also be invariant across domains (e.g. \cite{schumann2019transfer}). Besides, when it comes to in-processing methods, the optimisation problems for training the classifier need to re-defined so that they also account for distribution shift, to build models that are fair and accurate in the shifted deployment environment.

%Besides, in-processing methods aim to jointly optimise predictive accuracy and fairness; thus, when the distribution from the deployment environment is expected to be shifted from the source distribution, the objective function and constraints need to be redefined so that they are evaluated w.r.t. the distribution from the shifted deployment environment (i.e. the target population).  

The literature in the intersection between enforcing fairness and preserving predictive performance under distribution-shift has also studied the existence of possible trade-offs between both objectives, and several works have claimed that they might not be completely orthogonal. Building on that, recent research has explored under which conditions enforcing fairness implies robustness and vice versa. \citet{madras2018learning} empirically demonstrated that when fairness is adversarially learned following their method, the resulting representation is not only fair but can also transfer predictive performance as well as fairness guarantees to new tasks. \citet{maity2021does} showed that enforcing group notions of fairness mitigates algorithmic biases caused by sub-population shifts (e.g. under-representation of a given group) whenever imposing such constraint recovers the Bayes' classifier in the target domain. \citet{mukherjee2022domain} studied the similarity between enforcing individual fairness (IF) and domain adaptation and generalisation approaches, and derived the conditions under which enforcing IF achieves domain generalisation and vice-versa. Although the works above address specific scenarios where enforcing one criterion provides safe guarantees on the other, the scenarios they describe are too demanding and, therefore, hardly satisfied in real-world applications. In the light of such issue, their utility to preserve fairness guarantees and predictive performance in shifted environments is limited. Consequently, their existence does not call into question the necessity to develop methods to jointly address fairness concerns and distribution shift under more realistic conditions.

\section{Overcoming Distribution-Shift in Bias-Aware Scenarios}\label{s:main}

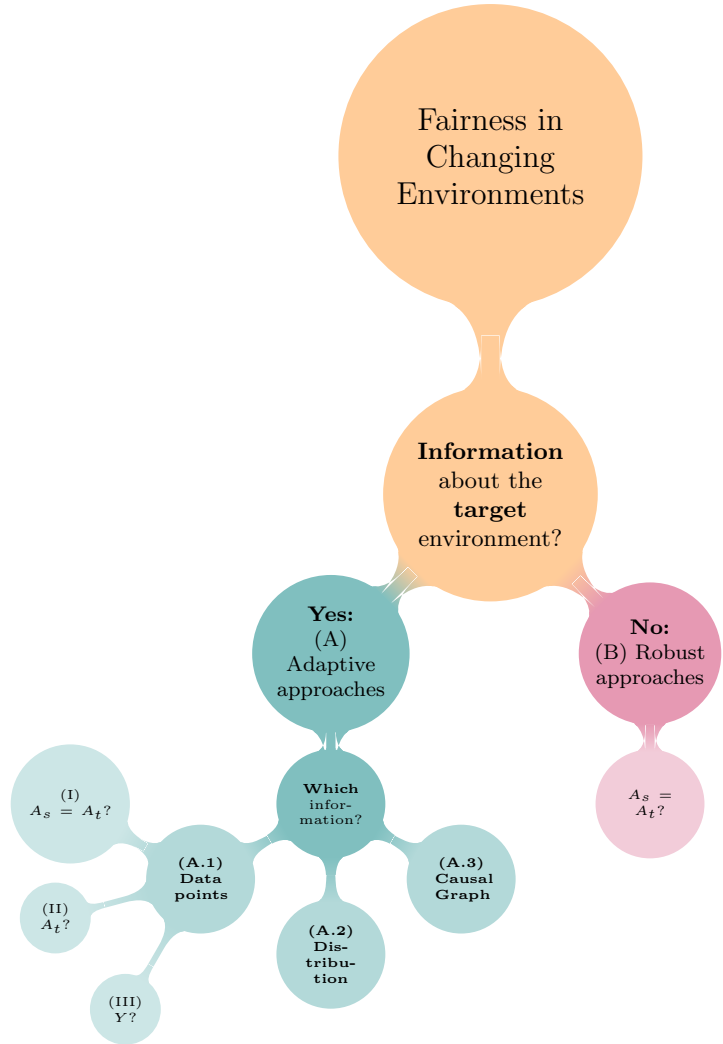
\begin{figure}
    \centering

\begin{tikzpicture}[mindmap, grow cyclic, every node/.style=concept, concept color=orange!40,
	level 1/.append style={level distance=4.5cm,sibling angle=40},
	level 2/.append style={level distance=3cm,sibling angle=90},
	level 3/.append style={level distance=2cm,sibling angle=90},
	level 4/.append style={level distance=2cm,sibling angle=60},
	level 5/.append style={level distance=2cm,sibling angle=45}
	]
	
	\node{Fairness in \\ Changing \\ Environments } [clockwise from=270]
child [concept color=orange!40] { node [minimum size=8em] {\textbf{Information} about the \textbf{target} environment?}
    [clockwise from=315]
	child [concept color=purple!40] { node [minimum size=4em] { \textbf{No:} \\ (B) Robust approaches}
	    [clockwise from=270]
	    child [concept color=purple!20] { node [minimum size=4em] {$A_s = A_t$? }}
	}
	child [concept color=teal!50] { node [minimum size=4em] { \textbf{Yes:} \\ (A) Adaptive approaches}
	    [clockwise from=270]
	    child [concept color=teal!50] { node [minimum size=4em] { \textbf{Which} information?}
	        [clockwise from=330]
            child [concept color=teal!30] { node [minimum size=4em] {\textbf{(A.3)\\ Causal Graph}}
	        }
	        child [concept color=teal!30] { node [minimum size=4em] {\textbf{(A.2) Distribution}}
	        }
	        child [concept color=teal!30] { node [minimum size=4em] {\textbf{(A.1) Data points}}
	            [clockwise from = 240]
	            child [concept color=teal!20] {node {(III) $Y$?}
	            }
	            child [concept color=teal!20] {node {(II) $A_t$?}
	            }
	            child [concept color=teal!20] {node [text width=4em] {(I) $A_s = A_t$?}
	            }
	        }
	    }
	}
};
	
\end{tikzpicture}

    \caption{Taxonomy-based classification of the methods that address distribution-shift under fairness constraints. The approaches are classified according to the assumptions they make regarding the information availability of target environment. First, we consider the two main scenarios, sketched by whether the target environment is known (\textbf{adaptive / robust} fair approaches). Inside the adaptive approaches, the methods are grouped based on the type of information they assume to be accessible from target environment, into \textbf{data-based}, \textbf{distribution-based} and \textbf{causal} fair adaptive approaches. Finally, inside the data-based approaches, we identify different scenarios grounding on the data-availability of the class label and the sensitive information, as well as the assumptions made on the latter ($Y$ available?, $A_t$ available?, assume $A_s = A_t?$).}
    \label{fig:methods}
\end{figure}

In this section we present the existing approaches that address robustness in performance while guaranteeing fairness in distribution-shift frameworks. Depending on the information available about the deployment environment, methods that overcome the adversities produced by distribution-shift have been developed in two different scenarios:
\begin{itemize}
    \item[(A)] If there is some information available from the target environment, the classifier initially trained on the source environment is \textbf{adapted} to the target. This is the case where, e.g. hospital A has an automated ML system to predict whether an individual suffers from a given disease, and it is to adapted to safely deploy it in hospital B, where the population is different. 
    \item[(B)] Otherwise, if there is no information about the target environment, methods are trained to be \textbf{robust} to whatever deployment distribution they might be applied in. This case arises when, e.g. the disease predicting tool from hospital A is aimed to be robust so that it is safely deployed in any possible hospital. 
\end{itemize}
Based on this distinction, the methods will be firstly classified into two main groups: \textbf{adaptive} methods and \textbf{robust} methods. The taxonomy proposed to characterize the different methods of this research field is summarized in Figure \ref{fig:methods}.

In the (A) cases, the kind of information known about the target environment is generally classified into three different types: data from the target population, the distribution from the target domain or the causal graph that jointly describes the data-generating processes from the source and target environments. The former situation gives rise to (A.1) \textbf{data-based} fair adaptive approaches and happens, e.g. when patient data is available in the target hospital. The second type  considers the scenario (A.2) in which there is certain knowledge regarding the \textbf{probability distribution} of the target environment, e.g. we (partially) know the distribution that describes the population of the target hospital. Finally, the third type requires (A.3) \textbf{causal} fair adaptive approaches, and can be found e.g. when accessing to instances from different hospitals is out of hand, but we instead have knowledge about which feature's distributions are expected to vary across different hospitals and which of them are expected to remain unaltered.   

Furthermore, in the (A.1) methods several different scenarios are possible depending on the assumptions about the class label and sensitive information of the target data. Note that in every scenario at least the non-sensitive features $X$ are known for the instances of the target environment.
In each scenario the data-based adaptation procedure will be different and will be characterized by three orthogonal aspects:
\begin{itemize}
    \item[(I)] \textbf{Is the sensitive feature assumed to be the same across domains ($A_s = A_t$)?} The sensitive information might vary from one domain to another. For example, the sensitive attribute available at the source domain (e.g. hospital A) might be gender, while that of the target domain (e.g. hospital B) being race. In these cases, the classifiers are developed aiming to satisfy fairness requirements w.r.t. gender in the source domain as well as w.r.t. race in the target domain.
    \item[(II)] \textbf{Are labels available on target data?} In the cases where the target data is available, the instances may consist of labelled (supervised adaptation) or unlabelled examples (unsupervised adaptation).  E.g. the information about the patients of the target hospital may contain the information about whether they actually suffer from the disease.
    \item[(III)] \textbf{Is the sensitive information available in the target domain?}  Similar to the previous consideration, the available data might lack from sensitive information ($A_t$). Indeed, the sensitive attribute of the patients of the target hospital (e.g. gender, race, age) might not be available.
  
\end{itemize}
Therefore, data-based fair adaptive approaches might face $2^3 = 8$ possible scenarios, that are detailed in Table \ref{tab:scenarios}, regarding the assumptions made on the class labels and sensitive information.

%The taxonomy proposed to characterize the different methods of this research field is summarized in Figure \ref{fig:methods}, where the different approaches are classified according to the assumptions they made regarding the target environment. In what follows, we introduce the reader to the main approaches that address covariate-shift for each of the likely deployment environment conditions:
%\begin{itemize}
 %   \item The \textbf{target} environment is \textbf{known}.
  %  \begin{itemize}
  %      \item \textbf{Instances} belonging to the target environment are available (8 possible scenarios).
 %       \item There is certain knowledge regarding the \textbf{probability distribution} of the target environment. 
 %       \item The \textbf{causal graph} that jointly describes the data-generating processes from the source and target environments is known.
  %  \end{itemize}
  %  \item The \textbf{target} environment is \textbf{unknown}.
%\end{itemize}

\subsection{Fair Adaptive Approaches}

The principal aim of fair adaptive approaches is to modify an existing accurate and fair classifier so that it can be safely deployed in an environment where the data generating process might differ from the original. This procedure is highly beneficial in three different scenarios. First, (A.1) when the target data is incomplete to build a new model from scratch either because there are few labelled examples, all instances are unlabelled or the sensitive information is lacking in the target environment. Further, it is as well an advantageous procedure when instances from the target environment are not available but some information is known either on the distribution (A.2) or on the causal graph (A.3). 

\subsubsection{Data-based fair adaptive approaches (A.1).}

These methods assume that some target data is accessible at training time and use it in combination to the labelled instances from the source domain in the training process. However, the datapoints from the source and target environments play different roles in the adaptation procedure, due to its distinct characteristics and information availability. Most of the methods are based on existing popular approaches that address distribution-shift in a fairness-agnostic scenario, modified to account for fairness guarantees in shifted scenarios (besides accurate performance). Depending on the information available from the target environment and the assumptions regarding the sensitive information, eight different scenarios are described in Table \ref{tab:scenarios}. Note that, the more assumptions the methods make regarding the available information from the target, the less flexible they become. Consequently, the most flexible adaptive method would be the one that grounds on scenario 8. On the other hand, methods that do not assume $A_s = A_t$, can always be employed in the equivalent cases where $A_s = A_t$ holds (e.g. methods that address scenario 5 can be used to address scenario 1).  

Among the existing data-based methods, the majority consider either scenarios 1 or 3, while there are even some that have not yet been covered by any method, specifically scenarios 2, 6 and 8. More precisely, most of the works assume there is no access to the target label (unsupervised adaptive methods), and among those that do assume its availability (supervised adaptive methods) the majority are grounded on the online learning setting where the data comes in a streaming fashion. Furthermore, the sensitive information is more often than not assumed to be available: indeed very few approaches contemplate such information might not be accessible in the target domain (e.g. \cite{coston2019fair}). Besides, the most popular assumption regarding the sensitive attribute is that it remains unchanged for the different domains, and therefore, those methods that assume that it may differ across domains are scarce ( e.g. \cite{schumann2019transfer, yoon2020joint}).
%CausalDA+RW \cite{singh2021fairness}, in scenario 5
In what follows, we describe the different approaches proposed to address the different scenarios in order of decreasing complexity, starting with those that assume $A_s = A_t$.
%, i.e. 1,5 and 7, . Then, we will follow with those approaches accounting for variations on the sensitive attribute (even-numbered scenarios) repeating the previous criteria based on the assumed amount of available information for training. 

\begin{table}[ht!]
    \centering
    \begin{tabular}{c|c|c|c}
        \multirow{2}{*}{\textbf{Scenario}} & \multicolumn{3}{c}{\textbf{Target information}}  \\
         & \textbf{$A_s = A_t$} & \textbf{$Y$ available} & \textbf{$A_t$ available} \\
        \hline
        1 & \cmark & \cmark & \cmark  \\
        \hline
        2 & \cmark & \cmark & \xmark \\
        \hline
        3 & \cmark & \xmark  & \cmark  \\
        \hline
        4 & \cmark & \xmark & \xmark  \\
        \hline 
        5 & \xmark & \cmark & \cmark \\
        \hline
        6 & \xmark & \cmark & \xmark \\
        \hline
        7 & \xmark & \xmark & \cmark \\
        \hline
        8 & \xmark  & \xmark & \xmark \\
          \hline
    \end{tabular}
    \caption{ Characterisation of the possible scenarios that data-based fair adaptive approaches may face regarding the availability of the class label and the sensitive information in target data, and the assumptions made on the sensitive attribute.}
    \label{tab:scenarios}
\end{table}

\textbf{(Scenario 1) Supervised adaptation with available sensitive information and $A_s=A_t$.} This scenario is most commonly addressed in \textit{online learning} settings, where the data is generated in a streaming fashion and its characteristics may evolve over time. Although it consists of a fully supervised setting, the label of an instance becomes available shortly after its arrival. In this framework, a prediction is made for every arriving instance using the current classifier, and the latter is updated after the ground truth labels of those instances become available. This setup is popularly named first-test-then-train or prequential evaluation. \citet{iosifidis2019fairness} proposed a pre-processing procedure: if the unfair behaviour of a classifier exceeds a threshold value $\epsilon$ for the set of instances arriving at time $t$, the chunk is modified either by massaging or by re-weighing the instances before feeding it to the base learner. The base learner can be any classifier that operates in an online learning setting. Post-processing methods can also be employed in conjunction with classical online learning classifiers. \citet{iosifidis2020fabboo} proposed one such method, applicable to decision boundary based online learning classifiers. They keep track of the cumulative fairness guarantees of the classifier, in order to avoid long term discriminatory effects that can go unnoticed when fairness concerns are taken into account for each particular time point. Whenever the cumulative discriminatory behaviour surpasses a used-defined threshold $\epsilon$ the decision boundary is adjusted to fulfil the fairness requirements. 

%Also different from previous approaches, they additionally address the class imbalance problem (related to the class label $Y$): they monitor the class imbalance of the incoming set of instances and weights the instances to ensure both classes are properly learned by the classifier.

%In particular, these approaches are based on an adaptive version of the Hoeffding Tree (HT) \cite{domingos2000mining, bifet2009adaptive}, a decision tree classifier popularly adopted in the scenario where the data arrives in streams, endowed with the ability to adapt to the changes on the underlying data generating process.

Besides, there are several in-processing approaches that adopt a \textit{decision-tree} based approach \cite{zhang2020feat, zhang2020flexible, zhang2021farf}. In-processing approaches that operate in an online learning setting adopt two different strategies w.r.t. shift detection and the consequent model adaptation: in all the cases, when distribution shift is detected the model is adapted to the new instances subject to fairness requirements; however, some approaches assume that the distribution has shifted only when a deterioration in the predictive performance is detected (e.g. \cite{zhang2020feat, zhang2020flexible}), while others account not only for performance deterioration but also for rising fairness violations for shift detection \cite{zhang2021farf}. For instance, among the first group of methods, \citet{zhang2020feat} proposed a new fairness-aware splitting criterion that reckons information and fairness gain. This criterion was further extended in \cite{zhang2020flexible} by adding a parameter to the definition to control the trade-off between information gain and fairness gain: at each time point, the practitioner can interpolate between a model that favours accuracy and a model that favours fairness. Besides, \citet{zhang2021farf} brought forward a fair and adaptive random forest (RF) that belongs to the second kind of approaches, that is, changes in the distribution are not only characterised in terms of performance deterioration, but also with increasing unfairness. Nonetheless, fairness is also accounted in the training and adaptation procedure: the base learners (i.e. decision trees) are chosen to increase the diversity of the ensemble w.r.t. discrimination; that is, they chose a diversified set of base classifiers, where each of them reflects a different discrimination representation. Besides, they under-sample the instances from the privileged group that have the positive (desired) class label to build fair base learners and, further, the predictions of the base learners are weighted according to their current fairness guarantees.

\textbf{(Scenario 3) Unsupervised adaptation with available sensitive information and $A_s=A_t$.} Most of the data-based adaptive works consider this scenario: they assume that target data is unlabelled but that there is access to the sensitive feature. Further, they suppose that the sensitive feature is that of the source domain. The first works that addressed this scenario adopted a reweighing approach \cite{kallus2018residual, coston2019fair}: they assign weights to the instances of the source domain to represent the population of the target domain, since the class labels are available only for the source data. Once these weights are known, it is possible to evaluate the expected value of any function on the target population using source samples. Depending on the fairness-enhancing strategy under consideration, the weighing function is employed and learned differently.
For example, \citet{kallus2018residual} proposed to employ the post-processing approach by  \citep{hardt2016} to enforce fairness on the target data. The fair method takes as input the $TPR$ and/or $FPR$ of a conventional ML classifier for the different sensitive groups, but evaluated on the target domain. Since the labels of the target instances are unknown, the $TPR$ and $FPR$ are estimated using the source data, by weighting the contribution of each instance by the weight function. In this approach, they use the classical covariate-shift instance weighing function, treating the sensitive attribute as a conventional feature:
\begin{equation}
    w(x,a) = \frac{\mathbb{P}_t(x,a)}{\mathbb{P}_s(x,a)}
    \label{eq_cov_a}
\end{equation}  
Therefore, in this case, the adaptation is exclusively made on the predictor derived on the post-processing step. On the other hand, \citet{coston2019fair} proposed a different framework considering an in-processing fairness-enhancing procedure, where fairness and distribution-shift are jointly considered at training time. Indeed, a new classifier is built for the target domain where the loss function is evaluated on the target population. Since the labels are inaccessible in the deployment environment, the loss function is estimated using the source data by weighing the contribution of each source instance: 
\begin{equation}
    \hat{\theta}  \equiv \argmin_{\theta \subset \Theta} \bigg [ \frac{1}{n} \sum_{i=1}^n w_i \mathcal{L} (\hat{y}(x_i, \theta), y_i) \bigg ]
    \label{eq_erm}
\end{equation}
where $\mathcal{L}$ represents the loss function between the predicted outcome $\hat{y}$ and the true label $y$. The weights $\{w_i\}_{i=1}^n$ are learned by means of an fairness-aware optimisation method, that aims to minimise a combination of a fairness loss ($\mathcal{L}_f$) and an approximated classification loss:
\begin{equation}
    \min_{w} \frac{1}{n} \sum_{i \in \mathcal{D}_s} w_{CS}(x_i) \mathcal{L} (\hat{y}(x_i, \hat{\theta}), y_i) + \lambda \mathcal{L}_f(\hat{y}(\cdot, \hat{\theta}))
    \label{wi}
\end{equation}
where $w_{CS}(x)$ is the conventional covariate-shift weighting function $w_{CS}(x)=\mathbb{P}_t(x)/\mathbb{P}_s(x)$ (similar to eq. \ref{eq_cov_a} but without the sensitive information). In particular, they consider the scenario in which the sensitive information is unavailable in the source domain. Therefore, the first term from eq. \ref{wi} estimates the classification loss on the target population by weighting the source population, where the labels are available; while the second term, evaluates the fairness loss on the target instances where the sensitive information is available. Therefore the proposed model goes as follows: for a given set of weights $w = \{w_i\}_{i=1}^n$ the optimal parameter values are learned (eq. \ref{eq_erm}), then the most representative set of weights is estimates solving eq. \ref{wi} using the optimal parameter choice obtained when solving eq. \ref{eq_erm}. This process is iteratively repeated until it converges to a stable solution. 
The main advantage of reweighing methods is that they are compatible with a diverse set of classification algorithms. However, it becomes problematic when the probability distributions of the source and target domain have distinct support, a scenario that is widely common in real-world settings.

%However, post-processing methods are known to have poor guarantees in terms of the balance between accuracy and fairness, and therefore, accounting for covariate-shift and fairness separately leads to suboptimal solutions. 

In other to overcome the weaknesses of instance weighing approaches, several authors have proposed to learn a new in-processing method for the deployment environment that jointly accounts for distribution shift and fairness by means of \textit{adversarial learning} \cite{goodfellow2014generative}: the adversary makes up the worst-case target distribution $\mathbb{Q}(x,a,y)$ that is coherent with the available target data, and the learner aims to find the parameter setting that optimises the predictive performance and fairness guarantees under such distribution. This can be seen as a minimax optimisation problem. Besides, note that, in a fairness-aware scenario, the worst case is defined not only by the predictive performance but also by fairness guarantees. Since fairness is achieved under the worst case, the model\textquotesingle s fairness is guaranteed on the test data. A variety of shifted scenarios have been proposed inside this framework where, for each case, $\mathbb{Q}(x,a,y)$ is defined differently. For instance, \citet{rezaei2021robust} considered a general covariate-shift scenario, and the worst-case distribution $\mathbb{Q}$ that approximates the target distribution is constrained to match a set of statistics of the source and target distributions. The constraints w.r.t. to the target data account for fairness requirements, and their characterisation varies for the different notions of fairness. \citet{du2021robust} adopted a similar approach but addressed the particular case when covariate-shift is a consequence of sample selection bias, i.e. when the source instances are a sub-population of the target environment that have been labelled under a biased labelling policy, which is unknown.  Thus, in this case, the adversary proposes the worst-case sample selection probability function. 

%Had the sample selection policy been known, any function could be evaluated on the target environment using the source samples. Unfortunately, the true sampling policy is actually unknown.

Furthermore, \citet{an2022transferring} proposed to incorporate an additional regularisation term to the objective function of in-processing fair methods, which aimed to transfer fairness to the target domain. In particular, they proposed a self-training framework where one model is training itself iteratively and fairness is transferred to the target domain using consistency training \cite{sohn2020fixmatch} on the data from both domains.

\textbf{(Scenario 4): Unsupervised adaptation without access sensitive information and $A_s = A_t$}. This scenario is similar to the previously addressed, but the main difference between them is that, in this case, there is no access to the sensitive information in the target domain: that is, only $X$ is known about the target data.  Consequently, the evaluation of fairness criteria on the target environment is not trivial. \citet{coston2019fair} proposed an in-processing reweighing method equivalent to that of the previous section (eq. \ref{eq_erm}), but with a different weight learning procedure, since the sensitive information is only available in the source domain. In fact, they aim to learn the weights that are as close as possible to the conventional covariate-shift weights (defined only by the non-sensitive attributes, available in both domains) subject to fairness constraints on the source data (where the sensitive information is available). That is, in this case the objective function reads as:
\begin{equation*}
    \min_{w} \sum_{i \in \mathcal{D}_s} |w_i - w_{CS}(x_i)| 
\end{equation*}
%subject to fairness constraints on prevalences of the source data:
\begin{equation*}
    s.t. \; \; \; \frac{\sum_{i \in \mathcal{D}_s; a_i = k ; y_i = 1} w_i}{\sum_{i \in \mathcal{D}_s; y_i = 1} w_i} \geq  \frac{\sum_{i \in \mathcal{D}_s; a_i = l ; y_i = 1} w_i}{\sum_{i \in \mathcal{D}_s; y_i = 1} w_i} - \delta \; \; \; \forall j,l \in A 
\end{equation*}

So far, the presented scenarios and the respective methods assume that the shifted environment contains the same sensitive information that the source domain, i.e. $A_s = A_t$. However, it might happen that the method needs to be adapted to a new environment where the sensitive information is different, i.e. $A_s \neq A_t$. In the following section we present those approaches that assume such adversity might arise in the deployment. In most cases, these scenarios are address by means of in-processing approaches by learning a new representation of the data that is invariant to the domain and sensitive information. Furthermore, this representation must be informative, i.e., the predictions made by a classifier when trained on top of this representation must be as accurate as possible. As we have been doing throughout, different scenarios will be identified according to the target information assumed to be available for adapting the classifier. 

\textbf{(Scenario 5) Supervised adaptation with available sensitive information and $A_s \neq A_t$}.

In this scenario, labelled data with sensitive information is available from the target environment, where the sensitive attribute is different from that of the source. This information might seem enough to build a new model for the deployment environment from scratch. However, the works that address this particular scenario focus on the case where target data itself is scarce and, therefore, insufficient to build a completely new model. Thus, to overcome such adversity, they propose a framework to benefit from a dataset belonging to a distinct population but having abundant labelled data to build a suitable model, even if this auxiliary dataset has a different sensitive attribute. For instance, in image processing tasks, developing countries tend to have scarce images due to limitations in their resources and, consequently, models behave poorly \cite{shankar2017}. These countries could benefit from images of wealthier countries with better imaging resources to build fair and accurate models, even if the populations differ notably and the sensitive information is different due to socioeconomic factors (e.g. in one country race might represent a critical issue, while in another country the discrimination is deemed w.r.t. gender). For that purpose, \citet{schumann2019transfer} proposed to learn a new representation of the data $h(x,a,y)$ and build a classifier $f$ that when trained on the new representation $h(x,a,y)$ of the data will achieve high predictive performance and fairness guarantees on the source and target domains, even when the sensitive attribute (and therefore the sub-population grouping) changes. For that purpose, they adopted an adversarial learning framework, where the objective function accounts for three losses: (1) classification loss when a classifier is trained on the new representation $h(x,a,y)$, and the losses that account for (2) the ability of the adversary to predict the sensitive attribute of the source data and (3) the ability of the adversary to predict which domain the instances belong to  when the new representation $h(x,a,y)$ is used.

%This scenario arises when we have labelled data from the source and target environments. Creating a model using both sources is beneficial if the data of one of both domains is insufficient to develop a model that behaves well in such environment, and besides, if the sensitive information is different in both approaches, 

\textbf{(Scenario 7) Unsupervised adaptation with available sensitive information and $A_s \neq A_t$}.

This scenario is similar to the previous but less restrictive: label information is assumed to be unavailable from the target data, and is therefore more difficult to address. In order to find a data representation that is invariant to both the domain and the sensitive information in such scenario, \citet{yoon2020joint} proposed an approach based on the Optimal Transport framework. With that, the new representation of the data is expected to minimise the dissimilarity between the distributions of: (1) the different sensitive groups in the source domain, (2) the privileged populations across domains and (3) the unprivileged populations across domains, and simultaneously train a classifier on such representation so that its predictions are accurate and fair in both domains. The distance between distributions is measured by the Wasserstein distance, which has many advantages over other existing divergence measures (such as, total variation distance, Kullback-Liebler divergence and Jensen-Shannon divergence) since it takes into account the underlying geometry and assigns a finite divergence value even if the two distributions do not share support.

\subsubsection{Distribution-based fair adaptive approaches (A.2).}

Some approaches consider the scenario in which there is knowledge about how the target distribution in shifted from the source distribution, i.e., they assume certain knowledge about $\mathbb{P}_t(x,a,y)$. For instance, \citet{giguere2021fairness} considered that the distribution shift is caused by \textit{demographic shift}, and assumed that the demographic proportions (i.e., $\mathbb{P}_t(a)$) are known in the target environment. They contemplate two degrees of knowledge: first, the case where the demographic proportions are known explicitly, and secondly, when such proportions are known to be bounded by some intervals. They built their model grounding on Seldonian algorithms \cite{thomas2019preventing}. They divided the training data into two blocks: the first block is used to train a fair algorithm; on the other hand, the data from second block is used to compute a high confidence upper bound of the fairness guarantees of the fair algorithm in the deployment environment. This upper bound incorporates the demographic information that is known about the target environment (exact proportions or bounds). If the fair algorithm is expected to behave fairly with high confidence on the target environment, their algorithms returns the initial fair model; otherwise, no model is provided. This framework can be employed with any procedure for training the fair classifier.

\subsubsection{Fair Causal Approaches (A.3).}

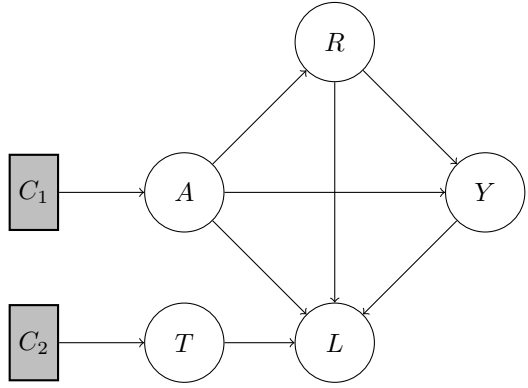
\begin{figure} 
    \centering

\begin{tikzpicture} [->, block/.style={draw, thick, minimum height=1cm, align=center}, elips/.style={ellipse,draw,minimum width=3em,minimum height=3em, align=center, inner ysep=0pt} ] 
  \node[block, fill = lightgray] at (0,0) (a) { $C_1$};  
  \node[elips] at (2,0) (b) {$A$};
  \node[elips] (c) at (4,2) {$R$};
  \node[elips] (d) at (6,0) {$Y$};
  \node[block, fill = lightgray] at (0,-2) (e) { $C_2$};
  \node[elips] at (2,-2) (f) {$T$};
  \node[elips] at (4,-2) (g) {$L$};

  \path[every node/.style={font=\sffamily\small}]
    (a) edge node [right] {} (b)
    (b) edge node [right] { } (c)
    (b) edge node [right] {} (d)
    (c) edge node [right] {} (d)
    (b) edge node [right] {} (g)
    (c) edge node [right] {} (g)
    (d) edge node [right] {} (g)
    (e) edge node [right] {} (f)
    (f) edge node [right] {} (g);

\end{tikzpicture}

    \caption{A joint causal graph that describes the influence of the distribution shift on the features. $C_1$ and $C_2$ are the context variables that describe the effect of distribution shift, i.e., the distributions of which variables are going to change because of the shift. $R$, $T$ and $L$ are non-sensitive features, $A$ is the sensitive feature and $Y$ the class label.}
    \label{fig:Causal}
\end{figure}

In general, causal approaches consider that there is no access to the target domain data, but presume the whole data generating process is described by means of a causal graph and the latter is known \cite{singh2019fair, singh2021fairness}.  Particularly, they use the \textit{joint causal inference} framework \cite{mooij2020joint}, a modelling approach that unifies the data generating process for all the distinct domains as a single causal graph that represents the underlying causal model. Figure \ref{fig:Causal} shows one such model. This framework introduces \textit{context variables} $C_i$ which describe how the different domains differ. In this way, they provide a complete overview about the plausible distribution shifts through the edges connecting the context variable(s) and the features. This information is highly relevant for the construction of classifiers that do not deteriorate under arbitrary shifts. Particularly, the key step of these causal approaches is to find a feature subset $S \subseteq \{X,A \}$ that satisfies the following assumptions: 
\begin{itemize}
    \item \textbf{Assumption 1: invariance of the classification error.} $S$ is a separating set, i.e. the conditional distribution of the class label $Y$ given the features $S$ is invariant across the different domains: $ Y \independent C \; | \; S $
    \item \textbf{Assumption 2: invariance of the fairness constraints.} Originally, \citet{singh2019fair} formulated one variant of this assumption for all fairness notions. However, they later claimed that the required condition changes with the fairness notion considered \cite{singh2021fairness}. 
\end{itemize}
These assumptions ensure that both the error and fairness constraints are identifiable in the target domain. With $S$ is identified, a classifier is built on top of it so that it does not deteriorate when deployed in any environment accounted in the causal graph:
\begin{equation}
    f(\theta^*) := \argmin_{\theta \in \Theta} \mathbb{E}_{(s,y) \sim P_t} \bigg [ \mathcal{L}(\theta, S, Y) \; : \; \textbf{g}(\theta, S, Y) \leq \epsilon \bigg ]
    \label{causalDA}
\end{equation}
where $\mathcal{L} : \Theta \times S \times Y \longrightarrow \mathbb{R}$ is the loss function and $\textbf{g}(\theta, S, Y)$ is the function denoting the fairness constraints. Thus this optimisation problem aims to find the parameter setting that minimises the classification error in the target domain while satisfying statistical fairness notions. 

The approaches based on causal frameworks suffer from several drawbacks: on the one hand, they require the knowledge of the causal graph that describes the data generating process; however, in general, the direct effects of the context variables on the rest of the features, as well as those between the features itself, are not trivial to define \cite{schrouff2022maintaining}. Further, these methods rely on strong shift assumptions which limits their deployment in real-world applications, where those premises are hardly hold. Consequently, in most cases, adopting these framework leads to solutions with no guarantees or trivial predictors.

\subsection{Fair Robust Approaches}

The above presented methods overcome the drawbacks caused by distribution-shift for particular deployment environments. However, those solutions can be highly sensitive to small perturbations, and an optimal solution can turn into a highly sub-optimal or even infeasible solution. Thus, it is essential to design fair robust approaches providing fair models that perform accurately on any possible unknown target environment and whose solutions remain nearly optimal under small perturbations of the data.  
%In this section the robust approaches are grouped according to how they incorporate the uncertainty regarding the deployment environment in the training procedure. 
Most of the fair robust approaches model the uncertainty over the deployment environment by means of an \textit{ambiguity} or \textit{uncertainty set} composed of perturbations of the empirical source distribution \cite{taskesen2020distributionally, du2021robust, mandal2020ensuring, du2021fairness, rezaei2020fairness}. In this framework,  known as \textit{distributionally robust optimisation} (DRO), models are made robust by finding the parameter setting that minimises the worst-case loss among all the probability distributions in the ambiguity set. Nonetheless, there are several approaches that follow distinct robustifying procedures \cite{wang2022robust, biswas2021ensuring}.

\subsubsection{Distributionally Robust Optimisation (DRO)}

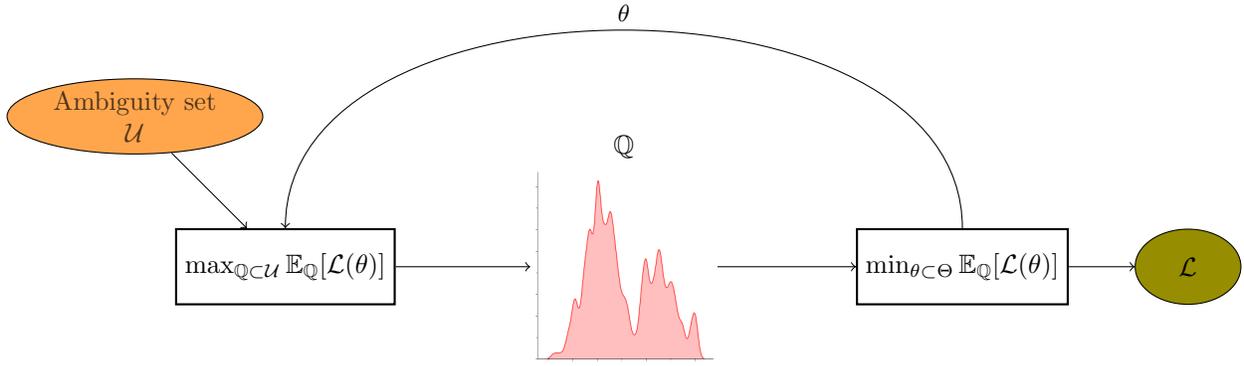
\begin{figure}
    \centering

\begin{tikzpicture} [->, block/.style={draw, thick, minimum height=1cm, align=center} ] 
  \node[elips, fill=orange, fill opacity=0.7] at (0,2) (a) { Ambiguity set \\ $\mathcal{U}$ };  
  \node[block] at (2,0) (b) {$ \max_{\mathbb{Q}\subset \mathcal{U}} \mathbb{E}_{\mathbb{Q}} [\mathcal{L}(\theta)]$};
  \node[inner sep=0pt] (c) at (6.5,0)
    {\includegraphics[width=.15\textwidth]{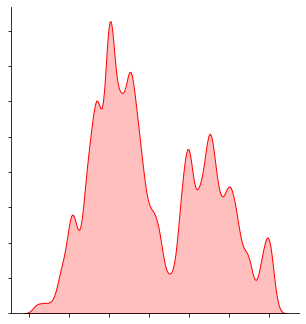}};
  \node[] (f) at (6.5,1.6) {$\mathbb{Q}$};
  \node[block] at (11,0) (d) {$ \min_{\theta \subset \Theta} \mathbb{E}_{\mathbb{Q}} [\mathcal{L}(\theta)]$};
  \node[elips, fill = olive] at (14,0) (e) {$\mathcal{L}$};

  \path[every node/.style={font=\sffamily\small}]
    (a) edge node [right] {} (b)
    (b) edge node [right] { } (c)
    (c) edge node [right] {} (d)
    (d) edge node [right] {} (e)
    (d) edge[bend right = 90] node [left, above, align = center] {$\theta$} (b);

\end{tikzpicture}

    \caption{ Scheme of the general workflow of fair DRO approaches. These approaches model the uncertainty regarding the deployment environment by means of an \textit{ambiguity} or \textit{uncertainty set} $\mathcal{U}$, and train the classifier to minimise the worst-case loss among all the probability distributions in $\mathcal{U}$.}
    \label{fig:DRO}
\end{figure}

These approaches assume that the possible deployment environments belong to an \textit{ambiguity} or \textit{uncertainty set} $\mathcal{U}$ and propose to train the models by means of an optimisation problem defined over the parameter setting that minimises the worst-case loss among all the probability distributions in $\mathcal{U}$. This is usually formulated as a minimax problem in which the loss in minimised with respect to the worst-case realisation of the perturbations of the ambiguity set: 
\begin{equation}
    \min_{\theta \in \Theta} \max_{\mathbb{Q} \subset \mathcal{U}} \mathbb{E}_{(x,a,y) \sim \mathbb{Q}} [ \mathcal{L}(\theta, X, A, Y) \; | \; \sup_{\mathbb{Q} \subset \mathcal{U}} g(\theta, X, A, Y) \leq 0]
\end{equation}
where $\mathcal{L} : \Theta \times X \times A \times Y \longrightarrow \mathbb{R}$ is a suitable loss function and $g(\theta, X, A, Y)$ are the functions defining the constraints (when applicable). Figure \ref{fig:DRO} shows a simplified scheme of DRO approaches. The following is a classification of the main DRO approaches according to the ambiguity set they consider to model distributional uncertainty, which are summed up in Table \ref{tab:DRO}. It is important to note that all of them assume that the sensitive attribute will remain unchanged for the possible unseen deployment environments.

%Besides, it is worth noting that the real probability distribution is unknown and is only observable by the nominal or empirical probability distribution. 

\textbf{Wasserstein ball.} \citet{taskesen2020distributionally} claimed the distribution describing the true data generating process is unknown and only indirectly observable through the training samples. They propose an ambiguity set that refers to the intersection between all the distributions within a Wasserstein ball centred at the empirical distribution and the set of all distributions under which the marginal of $(A,Y)$ matches the empirical marginal. The latter eliminates unrealistic data distributions from the ambiguity set. By taking the worst-case approach they propose to robustify a fair in-processing method (where fairness concerns are added as a penalisation in the objective function) against all the distributions in the set that, with high probability, will contain the true distribution. On the other hand, Du and Wu \cite{du2021robust} proposed a slightly different approach. They considered the special case where the distribution shift is caused by sample selection bias. Since the true sampling policy is unknown, they robustify the in-processing model w.r.t. to the worst possible sampling policy that could be encountered. That is, in this case, the uncertainty set contains the possible sampling policies. In particular, they cluster the data and assume that the instances belonging to the same cluster have the same sampling probability. Besides, the uncertainty set is a Wasserstein ball around the uniform selection ratio (that is, the case where all the clusters have the same selection probability). Moreover, different from the previous approach, fairness concerns are added as constraints that need to be satisfied in both the minimisation and the maximisation problems.

%propose a robust logistic regression (LR) model with an unfairness penalty defined by \textit{log probabilistic} notions of fairness. Indeed, the methods that rely on the empirical probability distribution are prone to overfitting and having poor out-of-sample performance. To overcome that, they propose to improve the robustness of a fair logistic regression model with respect to an ambiguity set they bring forward refers to the intersection between all the distributions within a Wasserstein ball centred at the empirical distribution and the set of all distributions under which the marginal of $(A,Y)$ matches the empirical marginal. 

%In this approach fairness is added as a constraint in the minimax problem and measured by decision boundary fairness \cite{zafar2017}. \\ 

\textbf{Weighted perturbations.} \citet{mandal2020ensuring} developed a double mini-max iterative algorithm where the ambiguity set is defined by weighted perturbations of the empirical training distribution. In each iteration, the weighted perturbation that maximises the loss is chosen first. Then, the algorithms aims to learn the parameter setting that minimises the weighted empirical risk of the given perturbation and is fair with respect to the worst possible fairness violation that can be found in the ambiguity set (and consequently, fair with respect to any perturbation in the ambiguity set). With this, they obtain a classifier that is fair and robust to any weighted perturbation of the training set. \citet{du2021fairness} also proposed an approach that is robust to instance-level weighted perturbations of the original dataset. They assume these weighted perturbations are linear combinations of some basis functions, e.g. Gaussian kernels. Fairness is enforced by means of additional constraints in both steps of the mini-max optimisation problem: the inner maximisation problem aims to find the perturbation that maximises the loss subject to fairness constraints, while the outer minimisation aims to find the parameter setting that minimises the worst-case loss subject to those constraints.  

%This method can be deployed with a wide range of fairness notions, such as DP and EO, and unfairness is measured by the weighted differences in performance metrics across sensitive groups. \\

%In their formulation they use the decision boundary fairness notion \cite{zafar2017} modified by the reweighing values, in order to obtain computational tractability. \\

\textbf{Statistics matching.} Rezaei \textit{et al.} \cite{rezaei2020fairness} propose to modify an in-processing method, where fairness guarantees are added as constraints, with the objective of achieving robust performance under covariate-shift. In this approach, the ambiguity set is composed by all the distributions that match the statistics of the training data, characterised by a vector valued feature function. This robust model is suitable for a wide range of fairness notions, such as, DP, EO and EOp. 

%However, for those definitions that include knowledge about the true label in their definition, an additional step is required to estimate the class label using the parametric distribution created from the chosen worst-case distribution.  

\begin{table}[ht!]
    \begin{tabular}{c|c|c|c|c}
        \textbf{Method} & \textbf{Ambiguity set} & \textbf{Loss} & \textbf{Fairness} & \textbf{Fairness criteria}  \\
        \hline
        \cite{taskesen2020distributionally} & Wasserstein ball centered & Log-loss & Weighted & log-probabilistic +   \\
         & at the empirical distribution & & penalty & $\{$ EO, EOp, .. $\}$ \\
         \hline
         \cite{du2021robust} & Wasserstein ball centered & Selection probability & Constraint & Decision boundary fairness \\
         & at uniform cluster selection ratio & weighted loss & & \cite{zafar2017}  \\
         \hline
         \cite{rezaei2020fairness} & Statistics matching with the & Log-loss & Constraint & EO, EOp,  \\
          & training data & & & DP, etc. \\
          \hline
          \cite{mandal2020ensuring} & Weighted perturbations of & Weighted & Weighted &  EO, EOp, \\
          & the training dataset & empirical risk & penalty & DP, etc. \\
          \hline
          \cite{du2021fairness} & Weighted perturbations of & Agnostic & Constraint & Decision boundary fairness \\
          & the training dataset & & & \cite{zafar2017} \\
          \hline
    \end{tabular}
    \caption{Characteristics of the most representative fair DRO approaches in terms of the defined ambiguity set, the loss function, how fairness is accounted for and the fairness notions that can be considered.}
    \label{tab:DRO}
\end{table}

\subsubsection{Deployment environment agnostic procedures.}

Although the DRO framework has been most popularly considered when building fair and robust models, there are also few works that do not make any assumptions about the unknown deployment environment. For instance, \citet{wang2022robust} aimed to build a fair model for which the curvature of the loss function is equivalent for all the sensitive groups. With that, the model will have similar robustness across the different sensitive groups, i.e. the generalisation capability of the model will be equivalent for all the data that is not supported in the source distribution, independent of the sensitive attribute. The base fair model is trained by means of a traditional fair adversarial training method to which they add another component in the loss function to account for the similarity of the loss curvature for the different sensitive groups measured by the Maximum Mean Discrepancy (MMD) \cite{gretton2012kernel}. On the other hand, \citet{biswas2021ensuring} proposed a robust model to address label shift (i.e., shift in the prevalence). In particular, they proposed to build decoupled classifiers for each sensitive group $a$. Each of those classifiers has two constituent blocks: (1) a quantifier to estimate the prevalence of a given dataset (with datapoints that belong to sensitive group $a$) and, (2) an ensemble classifier. Every base classifier of the ensemble is trained with a randomly selected subset of the original dataset (of instances belonging to $a$), where each of those subsets will have a different prevalence. Consequently, each of the base classifiers of the ensemble is specialised in providing accurate prediction in datasets with a given prevalence. Therefore, when the classifier is deployed in a unknown deployment environment, first the target data is divided into different sensitive groups. Then, for each sensitive group they estimate the prevalence using the quantifier, and finally, they use the base classifier of the ensemble of $a$ that has been trained with data having the most similar prevalence.

\section{Datasets and Benchmarking}\label{s:datasets}

%In this section we describe the datasets that are composed of data belonging to different domains, but preserving the functional form and containing sensitive information in at least one feature. Besides, we specify how to evaluate the model using such datasets.
%and highlight some observations by conducting an experiment with one of the datasets.

%\subsection{Datasets}

In the experimental section, many methods use popular datasets belonging to the literature of algorithmic fairness and create synthetic shifts to evaluate the proposed model. However, there are many datasets suitable for benchmarking purposes which naturally contain distribution shifts that arise in the real world as well as sensitive information. In what follows, we briefly describe the most popular datasets used in the literature of this research area.   

\textbf{American Community Survey (ACS) data \cite{ding2021}.} This dataset is a superset of the popular UCI Adult dataset constructed from the US Census publicly available data and it is one of the best suited collections. It covers different prediction tasks related to income, employment, health, transportation and housing. For each of the tasks they provide data belonging to all fifty US states and Puerto Rico, and five years of data collection between 2014 and 2018, both inclusive. That is, for each prediction task there are 255 distinct datasets available to account for real geographical and temporal variations. Further, it is also possible to concatenate each state's data to create datasets that enclose all the US. Besides, all the the tasks have sensitive information enfolded in the features for age, race, and sex. This structure enclosing multiple tasks and shift is collected in the Python package {\fontfamily{qcr}\selectfont folktables} \cite{folktables}. This set considers multiple prediction tasks, however,  the main ones are listed below:
\begin{itemize}
    \item \textbf{ACSIncome:} this task aims to predict whether an individual\textquotesingle s income exceeds the amount of 50,000\$. The instances of this datasets are individuals that are at least 16 years-old, have reported a minimum working activity of 1 hour per week the previous year and have an income of at least 100\$.
    \item \textbf{ACSPublicCoverage:} the objective of this task is to correctly predict whether an low-income individual is covered by public health insurance. The samples are individuals under the age of 65 whose income is lower than 30,000\$.
    \item \textbf{ACSMobility:} this prediction task focuses on whether an individual aged between 18 and 35 had the same residential address one year ago.
    \item \textbf{ACSEmployment:} aims to predict whether an individual aged between 16 and 90 is employed. 
    \item \textbf{ACSTravelTime: } to goal of this task is to predict whether the commute to work is higher than 20 minutes for individuals that are older than 16. 
\end{itemize} 
\textbf{WILDS.} WILDS \cite{koh2021wilds} is a collection of 10 datasets that describe a wide range of distribution shifts based on real-world applications, such as, shifts across hospitals for tumor identification. These datasets have three common characteristics: the shifts produce notable degradation on conventional ML classifiers that do not account for the shift, they reflect real-world settings and although they cannot be trivial, they must be solvable. Besides, WILDS contains a diverse set of tasks, data modalities, dataset sizes, and numbers of domains. This collection of datasets has been widely used in the field of fairness-agnostic distribution shift; however, several datasets contain sensitive information (e.g. CAMELYON17-WILDS, CIVILMODEL-WILDS, POVERTYMAP-WILDS, AMAZON-WILDS) thus are suitable for the analysis of classifier performance on a bias-aware distribution shift scenario. 

\textbf{MIMIC-III.} Medical Information Mart for Intensive Care III (MIMIC-III) is a open access critical care dataset \cite{johnson2016mimic, wang2020mimic}. It contains electronic health records (EHRs) of 53,423 patients (aged 16 or above) and 7870 neonates admitted to intensive care units (ICUs) at the Beth Israel Deaconess Medical Center hospital between the years 2001 and 2012 (both inclusive). Many works propose a cohort selection with patients aged  $>15$, and only consider the first ICU admission for each individual (e.g. \cite{wang2020mimic, purushotham2018benchmarking, johnson2017reproducibility, ghassemi2014unfolding, ghassemi2015multivariate, mcdermott2018semi, suresh2017clinical}). In particular, it collects data from 5 different ICUs: Medical ICU (MICU), Surgical ICU (SICU), Trauma Surgical ICU (TSICU), Cardiac Surgery Recovery Unit (CSRU) and Coronary Care Unit (CCU). The variables that characterize each patient include static features (e.g. age, admission type), time-series variables related to vital signs (e.g. heart rate, blood pressure) and laboratory test results (e.g. white blood cell counts). Further, it contains sensitive information such as age, race and gender. This dataset can be used for different tasks \cite{purushotham2018benchmarking}, including mortality prediction or estimation of the length of the stay at the ICU. Each of the ICUs can be viewed as a different domain or environment, since the population of patients varies from one ICU to others.

\section{Related Research Areas} \label{s:relatedfields}

In this section we describe other research areas that share some similarities with this field. Some of these areas enclose approaches that aim to preserve fairness guarantees in other non-standard albeit realistic scenarios. Besides, other fields are even more general and go beyond the conventional fairness-aware framework. Apart from those described below, we have seen that there are emerging connections, as yet little explored, but also worth mentioning such as robustness disparity in \cite{nanda2021fairness}.

\subsection{Fair inductive transfer learning}

Learning fair classifiers under distribution shift is a particular framework of \textit{fair transfer learning}: indeed, it constitutes a \textit{transductive} transfer learning scenario where although the data generating process changes across domains, the tasks are the same in both environments. Nonetheless, there are other transfer learning scenarios, such as, \textit{inductive} transfer learning, in which the tasks are different in the source and target domains. The main purpose of these frameworks is to benefit from combining previous knowledge about a collection of fairness-related tasks to obtain better and faster solutions, and requires access to labelled data. Using standard bias-agnostic transfer learning algorithms has proven to improve predictive accuracy on the target task at the cost of lowering the fairness guarantees of the predictions, a phenomenon called \textit{discriminatory transfer} \cite{lan2017discriminatory}. Nevertheless, preliminary work in \cite{zemel2013} suggest that fairness can indeed be transferable when is explicitly accounted for, and has been demonstrated to do so in recent work \cite{slack2020fairness, zhao2020fair, zhao2020primal, zhao2020unfairness, oneto2019, zhao2019rank}.

A popular fair inductive transfer learning setting is \textbf{fair meta-learning} \cite{slack2020fairness, zhao2020fair, zhao2020primal, zhao2020unfairness}, in which the main objective is to build a model that can be easily adapted to a new task with minimal data and iterations. In particular, most of the fair meta-learning approaches focus on the few-shot learning problem: these methods aim to learn general fair and accurate model (meta-model) valid for multiple tasks for initialisation, so that the classifiers for new tasks can be learned from minimal data with a small number of gradient update steps. Furthermore, \citet{zhao2021fairness} extended it to an online learning setting.

\textbf{Multi-task learning} is another inductive transfer learning framework. The key idea is to learn a set of relates tasks jointly to improve the predictive accuracy and fairness guarantees of the tasks as well as the generalisation guarantees. Although seemingly equivalent, the main purpose of the meta-learning and multi-task frameworks are indeed different: meta-learning aims to build a model that can be easily adapted to a new task with minimal data and iterations; on the other hand, multi-task learning aims to learn a shared representation to improve the predictive accuracy and fairness. \citet{oneto2020learning} proposed a multi-task learning framework to learn a representation that is fair as well as transferable.  Further, the representation learned under such setting is proven to generalise well to new unseen tasks with regards to predictive performance and fairness. Moreover, \citet{zhao2019rank} presented a fairness learning approach for multi-task regression models.

The fields of fair meta-learning and multi-task, as well as the fair methods that address distribution shift are concerned with the transfer of fairness guarantees. In the case of the former, the transfer is made across different but related tasks (where distribution shift can also happen); while for the latter, fairness guarantees are transferred across domains.

\subsection{Federated learning}

\begin{figure}[h]
\centering
\includegraphics[width=0.5\textwidth]{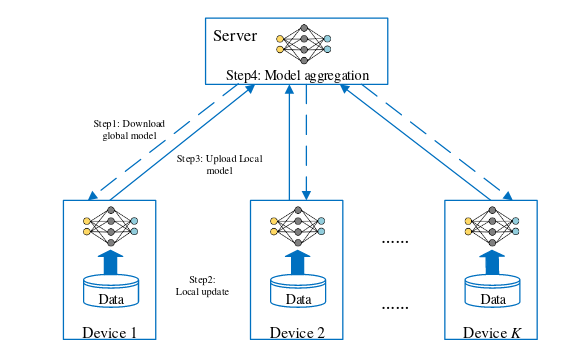}
\caption{Scheme representing the federated learning framework. Image from \cite{huang2020fairness}. The models are trained in a collaborative manner, where each of the data owners trains a local model on local data and shares the parameters of its model to generate a global model. }
\label{fig:federated}
\end{figure}

Federated learning (FL) \cite{McMMooRamHametal17} is a new ML framework to train models in a collaborative manner while protecting user privacy, data security, and government regulatory requirements. Each of the data owners trains a local model on local training data, and each of the owners shares the parameters of its model to generate a global model (see Figure \ref{fig:federated}). For example, automated vehicles process large amounts of information locally using ML models, while working cooperatively with other vehicles and computing centres. In practice, the local training data of each owner may follow a different distribution, causing the global model to have poor performance on testing data. Therefore, the distribution shift between the different data collections cannot be ignored when building the global model in order to achieve high accuracy and fairness guarantees when deployed on local data or previously unseen environments. Besides, the amount of data that is generated and collected varies considerably for every client (or device). Therefore, fairness plays an important role in FL \cite{du2021fairness, shi2021survey, zhou2021towards} not only related to data privacy or the discriminating behaviour of the model w.r.t. the sensitive information, but also in how the collaboration between different agents is made. 

Approaches grounded on federated learning need to address concerns related to distribution shift and fairness, albeit differently from those of the surveyed field: on the one hand, because model parameters are shared instead of the local data; and, on the other, because fairness concerns are not restricted to predictions, but encompass many other dimensions, such as, fairness in the collaboration when building the global model.

%With that, FL aims to build global models that follow a fair workflow and besides, guarantee high accuracy and fairness when deployed on local data or previously unseen environments. 

\subsection{Fairness-aware learning under corrupted data} 

In many real-world settings the training and validation data are oftentimes unreliable and biased, e.g. demographic data collected via surveys or online polls. Fair methods have shown to be vulnerable to data corruption \cite{mehrabi2021exacerbating} and, in the light of this issue, there is an emerging research line concerned about the development of fair classifiers that are robust to such adversity. Many works have been proposed to train fair models under noisy or adversarially perturbed sensitive attributes \cite{lamy2019noise, awasthi2020equalized, wang2020robust, celis2021fair, celis2021bfair, mehrotra2021mitigating}, labels \cite{olmin2022robustness, jiang2020identifying, wang2021fair, fogliato2020fairness} or both \cite{jo2022breaking}. Nonetheless, other works address more general data corruption scenarios \cite{roh2020fr, konstantinov2022fairness}. Preserving fairness guarantees under corrupted data is similar to doing so under distribution-shift since, for both scenarios, the training data is not representative of the real population from the deployment environment. However, the difference between the training and deployment environments is caused by distinct reasons: noisy, biased and unreliable measurements vs. changing environments.   

\subsection{Spurious correlations}

A spurious correlation is the dependence of a model\textquotesingle s predictions on aspects of the input data that should actually play a minor role, called nuisance factors. Those unwanted correlations are consequences of shortcut learning, which refers to the scenario in which algorithms rely on features that are easy to interpret and highly predictive of the class label. This association becomes problematic if the relation between the nuisance factor and the class label is unstable (i.e. varies for different deployment environments). In such case, whenever the model is deployed in an environment where the relation between the nuisance factors and the label is different (i.e. there exists distribution-shift), it performs even worse than random guessing \cite{goldstein2022learning}. For example, in an animal classification task using natural images, the background (e.g. grass, snow, water) can be highly predictive of the animal. As a consequence, the model, which apparently predicted cows accurately, failed considerably when tested on pictures where cows appeared outside the typical grass landscape \cite{beery2018recognition}. Several works have studied this difficulty and many approaches have been proposed to overcome it \cite{goldstein2022learning, geirhos2020shortcut, puli2021predictive, veitch2021counterfactual, makar2022causally}. Even if it is a general framework, it could be applied to a fairness-aware scenario, where sensitive attributes would constitute the nuisance factors. 

%\subsection{Dynamic modelling. Lifelong learning. Long term fairness. Dynamic fairness learning.} 
\subsection{Long term fairness} 
In online learning settings, where the deployment environment is constantly evolving, it is also crucial to be aware of the long-term behaviours of deployed ML algorithms and their potential consequences \cite{d2020fairness, liu2018delayed, hu2018short, mouzannar2019fair}. Several works have studied the long-term impact of conventional static fairness criteria on the well-being of the different sensitive groups \cite{d2020fairness, liu2018delayed, mouzannar2019fair}. In the light of the potential concerns that arise, others have proposed fairness-enhancing policies that ensure fairness/well-being in the long term \cite{hu2018short, ge2021towards}. The fair online learning approaches that have been considered in the survey do not account for long-term fairness concerns, but rather focus on satisfying fairness guarantees on the current population. Nonetheless, as an extension to those works, it would be beneficial to combine both fields and create adaptive methods that follow a training procedure that accounts for long-time fairness guarantees.

\section{Discussion and Future Research Challenges}\label{s:discussion}
%\paula{motivation}\\
ML research is facing increasingly complex problems. The task of obtaining reliable predictive algorithms, making use of statistical models, has been compounded by the challenge of dealing with changing environments. An additional difficulty arises from the fact that these algorithms operate by learning from potentially biased data, in which subgroups inevitably exist, so that ethical considerations must be taken into account to ensure equitable treatment. Initially, despite their different nature, both problems have been treated separately in the literature with similar methods providing satisfactory results. However, when trying to tackle both at the same time, the former no longer works and specific methods to address distribution-shift under fairness constraints are needed. Indeed, the continuous evolution intrinsic to our nature and society has given rise to the necessity of developing fair ML systems that preserve fairness guarantees when the data generating processes differ between the source (train) and target (test) domains. The reason for such a change in the environment, generally called distribution shift, can be very diverse and thus it is important first to have a classification of the different shifts considered in the literature so far, which was presented in section \ref{subs:changingenvironments}.

%\paula{applications}\\
The approaches that preserve fairness guarantees under distribution shift are highly relevant for many real-world applications involving geographical and temporal data. For instance, healthcare applications often face distribution shift scenarios, and for such critical settings, it is essential to provide robust fairness guarantees \cite{schrouff2022maintaining, chen2021probabilistic, wiens2014study, zech2018variable, finlayson2021clinician, chen2021algorithm}. Moreover, adaptive approaches could also overcome the disparities derived from the existing gap in imaging resources between countries \cite{shankar2017}.  Finally, it is worth noticing that there are many applications where the data gets easily outdated, and the underlying data generating process varies for future periods of time \cite{d2020fairness, zhang2020fair}. In order to avoid rebuilding the predictive model from scratch using the most recent labelled data, which is often expensive or impossible to recollect for such purposes, hard efforts can be avoided by adapting the existing model to the new environment. 

%Besides, it has been shown that computer vision models learn spurious age, gender, and race (to name some of the usual sensitive features) correlations when trained for seemingly unrelated tasks like activity recognition or image captioning.

% More precisely, these methods would be beneficial in many scenarios: (1) firstly, in hospitals with less resources that cannot afford a large labelling procedure could benefit from labelled data of other hospitals with better resources to create their own predictive models; (2) secondly, it is well known that data scarcity is one of the main issues in healthcare applications, which can be overcome by combining data from different hospitals (and, therefore, distributions) to improve the accuracy of the models; and (3) finally, although hospitals tend to create predictive models with their own patient\textquotesingle s data, they could increase their robustness to generalise well when they are deployed over the general population.

% Further, pedestrian detection models have sometimes shown trouble to recognise dark-skinned people \cite{wilson2019}, as a consequence of the abundance of light-skinned instances on the training data.

%\paula{main contribution}\\
This survey tried to contribute to the ML literature with the proposal of a novel taxonomy of the existing approaches that address robustness in performance while guaranteeing fairness in distribution-shift frameworks. Depending on whether there is information available about the deployment environment, the methods have been classified into two main groups: adaptive and robust methods. Among the adaptive family, we have identified three subgroups of methods depending on the type of information that is available about the target environment: data-based, distribution-based and causal approaches. Furthermore, we have proposed a reasonable and even more exhaustive classification of the existing data-based methods in the first subgroup, which has resulted in eight different scenarios depending on the assumptions about the class label and sensitive information of the target data. On the other hand, fair robust approaches are designed to preserve fairness guarantees when the deployment environment is completely unknown. The majority of these works follow a DRO-based approach and model the uncertainty over the deployment environment by means of an \textit{ambiguity set} composed of perturbations of the empirical source distribution. In addition to these, we have also found others that do not follow this specific framework and make no assumptions regarding the environment.

With the aim of giving a comprehensive overview on this line of research about preserving fairness guarantees in changing environments, this survey also reviewed in section \ref{s:datasets} the different datasets and benchmarking alternatives that have been used to implement previous methods, and pointed out the relation with other similar research fields in Section \ref{s:relatedfields}. As a result of these links on the one hand, and of the open problems that are left on the other, we are aware at this point that much remains to be done and identify in the following some future venues of research:  

\textbf{Theoretical guarantees.} Simultaneous to the proposal of methods to overcome model degradation in distribution-shift scenarios, several works study the theoretical guarantees of the proposed solutions in terms of the generalisation bound for the error and fairness guarantees. Indeed, those works define an upper-bound of the generalisation loss of either the error \cite{du2021robust, yoon2020joint}, the fairness guarantees \cite{schumann2019transfer, giguere2021fairness} or both \cite{mandal2020ensuring} characterised by the training loss and/or unfairness, model complexity and confidence. Besides, \citet{chen2022fairness} generalise the theoretical guarantees of fairness transferability for any ML algorithm. However, most of the approaches on the field do not provide theoretical guarantees. In the light of this void, we encourage the research community to include theoretical results in future approaches. Furthermore, the fairness transferability results obtained until now are limited to particular shifts or models; thus an extension of those results to more general scenarios would be an important future venue. 

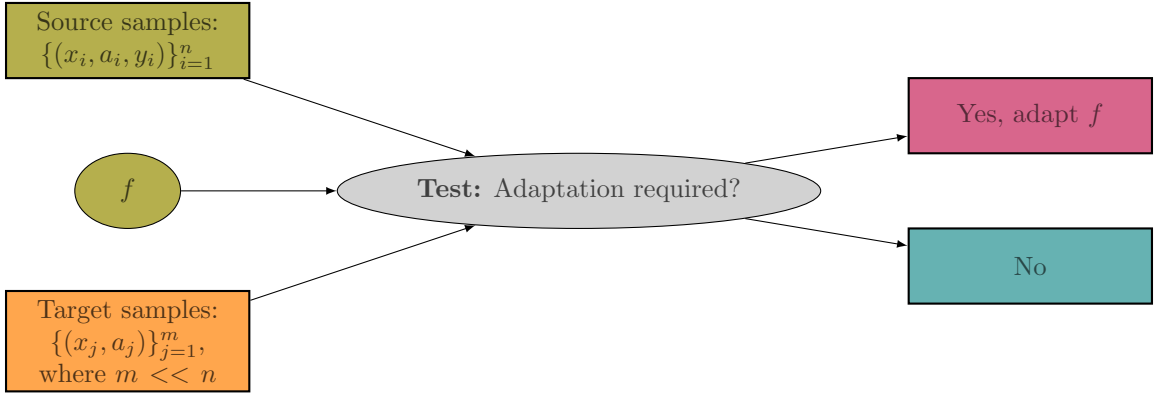
\begin{figure}
    \centering

\begin{tikzpicture}  
  \node[block, fill=olive, fill opacity=0.7] at (0,2) (a) {Source samples: $\{ (x_i, a_i, y_i) \}_{i=1}^n$ };  
  \node[elips, fill=olive, fill opacity=0.7] at (0,0) (b) {$f$};
  % you can use as many blocks by specifying the name and alphabets  
  \node[block, fill=orange, fill opacity=0.7] at (0,-2) (c) { Target samples: $\{ (x_j, a_j) \}_{j=1}^m$, where $m<<n$ };  
  \node[elips, fill=lightgray, fill opacity=0.7] at (6,0) (d) {\textbf{Test:} Adaptation required?};  
  \node[block, fill=purple, fill opacity=0.6] at (12, 1) (e) {Yes, adapt $f$ };  
  \node[block, fill=teal, fill opacity=0.6] at (12, -1) (f) {No};   
  % the yshift here specifies the shift in the position of the point on the y-axis. You can change the location according to the requirements.  
  % the value mentioned is the distance from (b) to (c). If the value is 0.5, then the block will be at the center of (b) and (c).  
  %\node[draw,inner xsep=4mm,inner ysep=7mm,fit=(a)(b),label={100:P}](g){}; % it is used to draw the box outside the combination of blocks. Here a block will be drawn outside the block (a) and (b). The xsep and ysep are the dimensions for the box. The label command here is used to set the location of the name of the outer A and B blocks.  
  %\node[draw,inner xsep=4mm,inner ysep=7mm,fit=(d)(e),label={80:Q}]{};  
    
  \draw[line] (a)-- (d);  
  \draw[line] (b)-- (d);  
  \draw[line] (c) -- (d); % you can change the location of the line to north, west, etc. depending on your requirements.  
  \draw[line] (d)-- (e); % -- signifies the line between the two blocks (c) and (d).  
 \draw[line] (d)-- (f);  
\end{tikzpicture}

    \caption{ Basic scheme of harmful shift detection techniques.}
    \label{fig:future1}
\end{figure}

\textbf{Harmful shift detection techniques.} One of the most important future venues is the design of harmful shift detection techniques (see Figure \ref{fig:future1}). Adaptive works assume knowledge about an existing shift between the training and deployment environments; however, little has been studied about how to detect the existence of a shift in practice. Besides, not all shifts will degrade the performance of the method in use. Therefore, it is particularly interesting to develop specific tests to detect when a classifier\textquotesingle s performance or fairness guarantees will not generalise well on the deployment environment. That is, when an adaptation is really necessary. Furthermore, those tests would be especially useful if the detection could be done by means of a small sample set of the target environment.  

%A preliminary work in ... identifies mean shifts that could lead to unfairness. (applicable to meta learning or any cov-shift? ) However, it only addresses a particular shift. We need more general methods.

\textbf{Uncertainty-aware evaluation framework of fair models.} It is well known that in real-world settings many assumptions made in the evaluation of the fair models do not hold. Therefore, a new evaluation framework is required for the safe deployment of the proposed methods in such settings. Assuming access to multiple independent datasets is non-realistic, and an exhaustive evaluation of a classifier in all the possible shifted deployment environments is infeasible. However, building on the idea of DRO approaches, it is possible to build a model-agnostic test where the classifier is tested under the worst-case target distribution (shifted from the source distribution) to verify whether the classifier is safe to be deployed in real-world settings without previous knowledge of the real environment it may encounter.

\begin{figure}
    \centering

    \begin{tikzpicture}  [->, block/.style={draw, thick, text width=2cm, minimum height=1cm, align=center}]
        \node[elips] at (0,0) (a) {$\mathbb{P}$};  
        \node[elips] at (-3,-2) (b) {$\mathbb{P}_1$};
        \node[elips] at (0,-2) (c) {$\mathbb{P}_2$};
        \node[elips] at (4,-2) (d) {$\mathbb{P}_k$};
        \node[font=\huge,align=center] at (2, -2) (e) {...};
        \node[block] at (-3,-4) (f) {$\{x_i, a_i, y_i \}_{i = 1}^{n_1}$};
        \node[block] at (0,-4) (g) {$\{x_i, a_i, y_i \}_{i = 1}^{n_2}$};
        \node[block] at (4,-4) (h) {$\{x_i, a_i, y_i \}_{i = 1}^{n_k}$};
        \node[] at (8,-2) (i) {$\mathbb{P}_i(x,a,y) \neq \mathbb{P}_j(x,a,y) \; \; \forall i \neq j$};
        
        \path[every node/.style={font=\sffamily\small}]
            (a) edge node [right] {} (b)
            (a) edge node [right] {} (c)
            (a) edge node [right] {} (d)
            (b) edge node [right] {} (f)
            (c) edge node [right] {} (g)
            (d) edge node [right] {} (h);
  
    \end{tikzpicture}
    \caption{ Multi-source setting.}
    \label{fig:multi}
\end{figure}
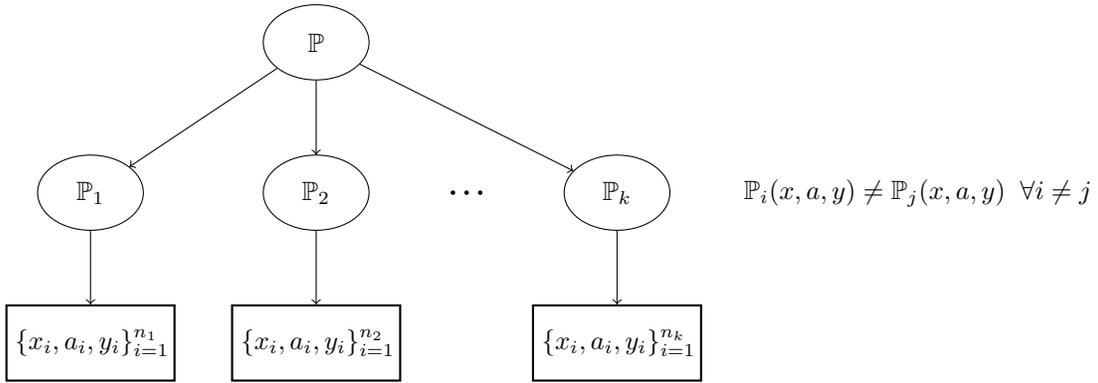

\textbf{Multi-source approaches.} The works that are part of the research area surveyed in this paper consider a single source domain. However, in many real-world cases one might have access to labelled instances belonging to different underlying data distributions that share the functional form (see Figure \ref{fig:multi}). Thus building models that account for a multi-source setting is of primary interest, either if they are adaptive or robust approaches.

\textbf{Fair robust approaches that account for $A_s \neq A_t$}. Up to the author\textquotesingle s knowledge, all the robust methods assume that the sensitive attribute will be the same in any possible deployment environment, which limits the applicability of the methods. With that, it is important to develop robust methods that account for the possibility of variation on the sensitive attribute in the likely deployment environments. 

\textbf{Multi-dimensional sensitive attribute}. All the methods proposed in the field assume that each instance is defined by a single sensitive attribute. Besides, in most cases, the sensitive attribute is presumed to have two possible values. However, the real-world settings in which the models are deployed are far more complex: individuals contain many overlapping sensitive attributes, such as, race, gender, sexual orientation, class, and disability, and are therefore affected by a combination of systems of power and oppression \cite{foulds2020bayesian}. Consequently, the fair methods that account for distribution shift should go beyond the binary sensitive attribute, and further consider multi-dimensional sensitive attributes.  

\section*{Acknowledgements}

This research was supported by a European Research Council (ERC) Starting Grant for the project “Bayesian Models and Algorithms for Fairness and Transparency”, funded under the European Union’s Horizon 2020 Framework Programme (grant agreement no. 851538); by the Basque Government through the BERC 2022-2025 program and by the Spanish Ministry of Science, Innovation, and Universities (BCAM Severo Ochoa accreditation SEV-2017-0718).

\addcontentsline{toc}{section}{References}

\printbibliography

\end{document}